\documentclass{article}

\usepackage{microtype}
\usepackage{graphicx}
\usepackage{subfigure}
\usepackage{booktabs} 

\usepackage{hyperref}

\usepackage[accepted]{icml2023}

\usepackage{amsmath}
\usepackage{amssymb}
\usepackage{mathtools}
\usepackage{amsthm}

\usepackage[capitalize,noabbrev]{cleveref}

\theoremstyle{plain}

\theoremstyle{definition}

\theoremstyle{remark}

\usepackage[textsize=tiny]{todonotes}

\usepackage{enumitem}
\usepackage{mathtools}
\newcommand{\Mat}{\boldsymbol}
\newcommand{\Set}{\mathcal}

\newcommand{\real}{\mathbb{R}}

\usepackage{amsmath}

\usepackage{graphicx}
\usepackage{amsmath}
\usepackage{amssymb}
\usepackage{booktabs}
\usepackage{makecell}
\usepackage{multicol}
\usepackage{multirow}
\usepackage{color}
\usepackage{xcolor}
\usepackage{colortbl,booktabs}

\icmltitlerunning{Graph Ladling: Shockingly Simple Parallel GNN Training without Intermediate Communication}

\begin{document}

\twocolumn[
\icmltitle{Graph Ladling: Shockingly Simple Parallel GNN Training \\without Intermediate Communication}



\icmlsetsymbol{equal}{*}

\begin{icmlauthorlist}
\icmlauthor{Ajay Jaiswal}{yyy}
\icmlauthor{Shiwei Liu}{yyy}
\icmlauthor{Tianlong Chen}{yyy}
\icmlauthor{Ying Ding}{yyy}
\icmlauthor{Zhangyang Wang}{yyy}

\end{icmlauthorlist}

\icmlaffiliation{yyy}{University of Texas at Austin}

\icmlcorrespondingauthor{Ajay Jaiswal}{ajayjaiswal@utexas.edu}

\icmlkeywords{Machine Learning, ICML}

\vskip 0.3in
]



\printAffiliationsAndNotice{\icmlEqualContribution} 

\begin{abstract}
Graphs are omnipresent and GNNs are a powerful family of neural networks for learning over graphs. Despite their popularity, scaling GNNs either by deepening or widening suffers from prevalent issues of \textit{unhealthy gradients, over-smoothening, information squashing}, which often lead to sub-standard performance. In this work, we are interested in exploring a principled way to scale GNNs capacity without deepening or widening, which can improve its performance across multiple small and large graphs. Motivated by the recent intriguing phenomenon of model soups, which suggest that fine-tuned weights of multiple large-language pre-trained models can be merged to a better minima, we argue to exploit the fundamentals of model soups to mitigate the aforementioned issues of memory bottleneck and trainability during GNNs scaling. More specifically, we propose not to deepen or widen current GNNs, but instead present \textbf{a data-centric perspective} of model soups tailored for GNNs, i.e., to build powerful GNNs. By dividing giant graph data, we build multiple independently and parallelly trained weaker GNNs (soup ingredient) without any intermediate communication, and \textit{combine their strength} using a greedy interpolation soup procedure to achieve state-of-the-art performance. Compared to concurrent distributed GNN training works such as \cite{zhu2023simplifying}, we train each soup ingredient by sampling different subgraphs per epoch and their respective sub-models are merged only after being fully trained (rather than intermediately so).
Moreover, we provide a wide variety of model soup preparation techniques by leveraging state-of-the-art graph sampling and graph partitioning approaches that can handle large graphs. Extensive experiments across many real-world small and large graphs, illustrate the effectiveness of our approach and point towards a promising orthogonal direction for GNN scaling. Codes are available at:  \url{https://github.com/VITA-Group/graph_ladling}.

\end{abstract}

\vspace{-0.5em}
\section{Introduction}

Graphs represent a myriad of real-world data from social networks, knowledge graphs, gene expression networks, etc. Graph neural networks (GNNs)  \cite{Kipf2017SemiSupervisedCW,defferrard2016convolutional,velivckovic2017graph,You2020L2GCNLA,Gao2018LargeScaleLG,chiang2019cluster,zheng2021cold,chen2018fastgcn,duan2022a,thekumparampil2018attention}, which use message passing (MP) strategy at their core for aggregating knowledge from neighbors, have been widely accepted as powerful algorithmic tools for learning over graphs. Although message passing provides GNNs superior performance over traditional MLPs, the nature of evolving massive topological structures prevents MP-based GNNs from scaling to industrial-grade graph applications, and the majority of state-of-the-art GNNs are only tested on small graph datasets. Additionally, due to the prevalent issues such as unhealthy gradients, over-smoothening and squashing \cite{li2018deeper,NT2019RevisitingGN,Alon2021OnTB,jaiswalold,liu2021overcoming} while training GNNs, increasing model capacity either by deepening (stacking more layers) or widening (increasing neighborhood coverage) often lead to sub-standard performance.

Previously, conforming to the empirical scaling laws \cite{kaplan2020scaling}, where the final model quality has been found to have a power-law relationship with the amount of data, model size, and compute time; several works \cite{li2021training,jaiswalold,zhou2021dirichlet} have attempted to scale GNNs (up to 1000 layers) assuming that processing larger graphs would likely benefit from more parameters. Unlike conventional deep neural networks, exploiting scale to revamp information absorption is not straight-forward for GNNs, and numerous existing works rely on architectural changes, regularization \& normalization, better initialization \cite{Li2019CanGG,Chen2020SimpleAD,li2018deeper,Liu2020TowardsDG,Rong2020DropEdgeTD,Huang2020TacklingOF,Zhao2020PairNormTO,Zhou2021UnderstandingAR,jaiswalold} for improving the trainability and try to overcome astonishingly high memory footprints by mini-batch training, i.e. sampling a smaller set of nodes or partitioning large graphs \cite{hamilton2017inductive,chen2018fastgcn,Zou2019LayerDependentIS,chiang2019cluster,zeng2019graphsaint}. While these methods are a step in the right direction, they do not scale well as the models become deeper or wider, since memory consumption is still dependent on the number of layers. 
We are interested in exploring an orthogonal step: \textit{Does there exist a principled way to scale GNNs capacity without deepening or widening, which can improve its performance across small and large graphs?  }

Recently, for large pre-trained models with many applications in computer vision \cite{Han2020ASO,li2023cancergpt,Mao2022SingleFA,jaiswal2021scalp,Zheng2021EndtoEndOD} and natural language processing \cite{talmor2018commonsenseqa,Jaiswal2021RadBERTCLFC,zheng2023outline,liu2023sparsity,chen2023sparse,jaiswal2023attend}, several works \cite{wortsman2022model,ilharco2022patching,juneja2022linear} investigate  the intriguing phenomenon of ``model soup'', and have shown that weights of multiple dense fine-tuned models (candidate \underline{ingredients} of soup) can be merged together into better generalizable solutions lying in low error basins. Despite enormous attention in NLP, it is unexplored for GNNs, presumably due to traditional wisdom that unlike large pre-trained transformers in NLP, current state-of-the-art GNN's model capacity is under-parameterized apropos of gigantic graphs.  Despite some recent works \cite{WANG2022346,math10081300} illustrating the benefits of GNNs ensembling, they exhibit high computational cost during inference which worsens in the context of large graphs. Motivated by the mergability of multiple fine-tuned models illustrated by model soups, in this work, we raise the research question: \textit{Is it possible to leverage the fundamentals of model soups to handle the aforementioned issues of memory bottleneck and trainability, during scaling of GNNs?}

To this end, we propose not to deepen or widen current GNNs, but instead explore a data-centric perspective of dividing ginormous graph data to build independently and parallelly trained multiple comparatively weaker models without any intermediate communication of model weights, and merge them together using  \textit{greedy interpolation soup} procedure to achieve state-of-the-art performance. Our work draws motivation from recent advancements in parallel training of \textit{pretrained language models} (LMs). For example, Branch-Merge-Train (BTM) \cite{Li2022BranchTrainMergeEP} learns a set of independent expert LMs specializing in different domains followed by averaging to generalize to multiple domains, and Lo-Fi \cite{wortsman2022fi} illustrates the futility of communication overhead in data-parallel multi-node finetuning of LMs. Although these techniques seem to work for \textit{large pre-trained LMs}, it is still unclear and unexplored if they will work for comparatively much \textbf{smaller GNNs in the  training-from-scratch regime}. Moreover, GNNs deal with graph-structured relational data unlike independent samples in LMs and have their own unique set of challenges in their trainability, which makes it interesting to understand if soup phenomenon will help or deteriorate GNNs performance.

To our surprise, we found that independently trained GNNs from scratch can be smoothly aggregated using our greedy soup interpolation procedure to create a better generalizable GNN that performs exceptionally well. It suggests that linear scaling of GNNs either by deepening or widening is not necessarily the right approach towards building high-quality generalizable GNNs, and model soups can be an alternative.
Note that, unlike the conventional model soup, we explore greedy weight interpolation for graph models, and in a well-motivated \textbf{data-centric} perspective (where it matters the most) considering the exploding size of real-world graphs. More specifically, in our work, we \underline{firstly} illustrate easy adaptability of model soups across multiple SOTA GNN architectures trained on multiple small and large graphs (unexplored till now) and \underline{secondly} present a novel data-centric perspective of model soups for large graph-structured relational data within constraints of available computational resource. Moreover, we extend current state-of-the-art graph sampling and partitioning techniques to facilitate the training of candidate soup ingredients which can be seamlessly combined at end for better generalization. We also compare our recipe with distributed GNN training, e.g., concurrent work \cite{zhu2023simplifying}, in Section 4.1.

Our primary contributions can be summarized as:
\vspace{-0.5em}
\begin{itemize}
    \item [$\blacksquare$] We illustrate the \textit{harmonious adaptability} of model soups for graph-structured data and experimentally validate its performance benefits across multiple GNN architectures and graph scales. Our experiments reveal \textit{orthogonal knowledge} stored in the candidate models which can be surprisingly aggregated during soup preparation using our greedy interpolation procedure. 
    
    \item [$\blacksquare$] We present a novel \textit{data-centric perspective} of model soups tailored for GNNs and carefully study the benefits of independent and parallel training of candidate models and their mergability in scenarios without the luxury of having computation resources to process entire graph, by extending state-of-the-art (SOTA) graph sampling and partitioning algorithms. 

    \item [$\blacksquare$] Our extensive experiments across multiple large-scale and small-scale graphs \texttt{\{Cora, Citeseer, PubMed, Flickr, Reddit, OGBN-ArXiv, OGBN-products\}} using multiple GNN architectures \texttt{\{GCN, JKNet, DAGNN, APPNP, SGC, GraphSAGE, CluterGCN, GraphSAINT\}} validates the effectiveness of our approach.
\end{itemize}

\section{Methodology}
\subsection{Preliminaries} 
Graph Neural Networks (GNNs) exploit an iterative message passing (MP) scheme to capture the structural information within nodes’ neighborhoods. Let $G = (\Set{V}, \Set{E})$ be a simple and undirected graph with vertex set $\Set{V}$, edge set $\Set{E}$ and .Let $\Mat{A} \in \real^{N \times N}$ be the associated adjacency matrix, such that $\Mat{A}_{ij} = 1$ if $(i, j) \in \Set{E}$, and $\Mat{A}_{ij} = 0$ otherwise. $N = \lvert \Set{V} \rvert$ is the number of nodes. Let $\Mat{X} \in \real^{N \times P_{0}}$ be the node feature matrix, whose $i$-th row represents a $P_{0}$-dimension feature vector for the $i$-th node. To facilitate understanding, a unified formulation of MP with K layers is presented as follows:

\vspace{-0.6cm}
\begin{multline}
    \Mat{X}^{\Mat{(K)}} = \Mat{A}^{\Mat{(K-1)}}\sigma(\Mat{A}^{\Mat{(K-2)}}\sigma(\cdot \cdot \cdot\sigma(\Mat{A}^{(0)}\Mat{X}^{(0)} \\ \Mat{W}^{(0)}  ))\cdot \cdot \cdot)\Mat{W}^{\Mat{(K-2)}})\Mat{W}^{\Mat{(K-1)}}
    \label{eq:messagepassing}
\end{multline}
\vspace{-0.6cm}

where $\sigma$ is an activation function (e.g. ReLU), $\Mat{A}^{(l)}$ is the adjacency matrix at $l$-th layer, and $\Mat{W}^{(l)}$ is the feature transformation weight matrix at $l$-th layer which will be learnt for the downstream tasks.

\begin{figure}[h]
    \centering
    \includegraphics[width=\linewidth]{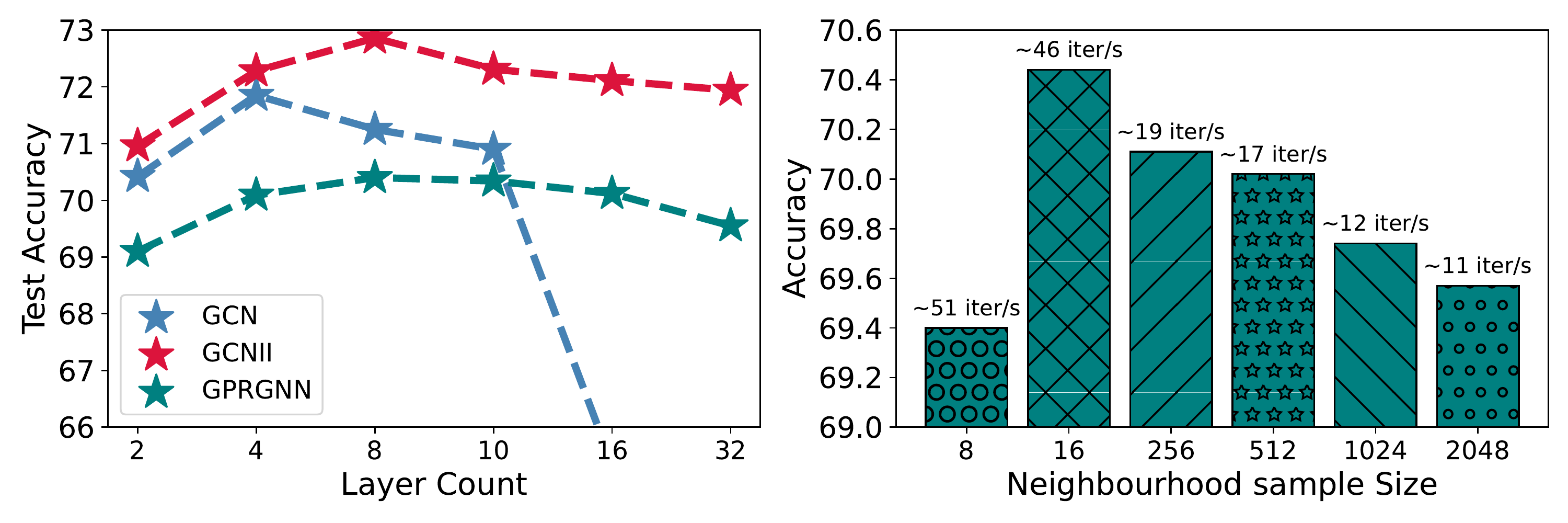}
    \vspace{-1.8em}
    \caption{(Left) Performance comparison of GCN, GCNII, and GPRGNN on \texttt{OGBN-arxiv} dataset with increasing layer count. (Right) Performance comparison of 2-layer GraphSAGE on \texttt{OGBN-arxiv} dataset with increasing neighbor sampling threshold. Results illustrate that deepening and widening the model capacity doesn't necessarily help in improved performance.}
    \label{fig:motivation}
\end{figure}

\subsection{GNNs and Issues with Model Explosion}
Despite the enormous success of GNNs in learning high-quality node representations, many state-of-the-art GNNs are shallow due to notorious challenges in their trainability, resulting in sub-standard performance. Recently, we have been observing a massive explosion in the topological structure of real-world graph data, which raises demand to increase the model capacity to facilitate capturing long-range dependencies, and relaxing the widely-adopted restrictions to learn from limited neighbors subjected to resource constraints due to the design issues of message passing strategy. 

Several works \cite{li2021training,jaiswalold,zhou2021dirichlet} have attempted the empirical scaling law from NLP, which suggests that over-parametrized models have better generalization capabilities, and proposed to similarly scale GNNs by stacking layers citing their increased ability to capture long-term dependencies. However, scaling GNNs capacity either by deepening to capture long-range dependencies or by widening to increase neighborhood aggregation is inhibited by the prevalent issues of unhealthy gradients during back-propagation leading to poor trainability, over-smoothening causing feature collapse, and information squashing due to overload and distortion of messages being propagated from distant nodes.   

Figure \ref{fig:motivation} illustrates the effect of state-of-the-art GNNs scaling from the perspective of deepening and widening on \texttt{OGBN-arxiv}. It can be clearly observed that vanilla-GCN, GCNII and GPRGNN, all suffer significant performance from drop  with increasing model capacity. Note that this substandard performance still exists in GCNII and GPRGNN, which are equipped with SOTA architectural modifications (eg. skip-connections), regularization, and normalization. Additionally, as shown in Figure \ref{fig:motivation}(right), increasing the breadth of message passing to capture more and more neighborhood information for GraphSAGE on \texttt{OGBN-arxiv} does not necessarily help in improving the performance rather increases the memory footprints and reduce throughput (iterations/sec). 

To this end, we argue that model explosion is not necessarily the only solution towards building high-quality generalizable GNNs and propose an orthogonal direction inspired by the recent distributed and parallel training platform of model soups. We propose to build independently and parallelly trained multiple comparatively weaker models without any intermediate communication, and merge their knowledge to create a better generalizable GNN. In the next sections, we will explain how model soups can be easily adapted for many state-of-the-art GNNs, and how can we smoothly extend the current graph sampling and partitioning to create candidate ingredients for the model soup.  

\begin{algorithm}[tbh]
   \caption{Greedy Interpolation Soup Procedure}
   \label{alg:example}
\begin{algorithmic}
   \STATE {\bfseries Input:} Soup Ingredients $\Mat{M} = \{M_{1}, M_{2}, ..., M_{K}\}$
   \STATE $\Mat{M}_{sorted} \leftarrow \texttt{SORT}_{ValAcc}(\Mat{M})$
   \STATE soup $\leftarrow \Mat{M}_{sorted}[0]$
    \FOR{$i=1$ {\bfseries to} $K$}
        \FOR{$\alpha$ {\bfseries in} $\texttt{linspace}(0,1,step)$}
            \IF{\texttt{ValAcc}(\texttt{interpolate}(soup, $M_i$, $\alpha$)) $\geq$ \texttt{ValAcc}(soup)}
                \STATE soup $\leftarrow$ \texttt{interpolate}(soup, $M_i$, $\alpha$)
            \ENDIF
        \ENDFOR
    \ENDFOR
\end{algorithmic}
\label{alg:souping}
\end{algorithm}

\subsection{Model Soups and current state-of-the-art GNNs}
In this section, we discuss about the harmonious adaptation of model soups for various state-of-the-art GNNs. Although model soups have been recently receiving ample attention for parallel training of large language pre-trained models, it is still unclear and unexplored if they will work for comparatively much smaller GNNs trained from scratch. We for the first time observed that unlike using a pre-trained initialization in LLMs, GNNs optimized independently from the \underline{\textbf{same random initialization}} organically lie in the same basin of the error landscape, and they can be smoothly aggregated using linear interpolation across their weight parameters, with significant performance gain.

More specifically, we propose to 
create $\Mat{K}$ instances (soup ingredients) of the same GNN architecture of interest, sharing a common random initialization to ensure that the optimization trajectory does not deviate significantly during training facilitating smooth mergability. Subject to resource availability, GNN instances are trained independently in isolation across multiple GPUs using the hyperparameter configuration set ${\{h_{1}, h_{2}, ..., h_{K}\}}$ designed with slight variations in the learning rate, weight decay, seed values, etc. 

Once the soup ingredients are ready, we start preparing the soup using our greedy interpolation soup procedure as illustrated in Algorithm \ref{alg:souping}, which in general follows \cite{wortsman2022model}. The greedy soup sequentially adds each model as a potential ingredient in the soup, and only keeps the model in the soup if it leads to improving performance on the validation set. For merging two ingredients, we perform a search for an optimal interpolation ratio $\alpha \in [0, 1]$ that helps in performance gain, else the ingredient is discarded. 

In comparison with several state-of-the-art GNNs, we experimentally illustrate that GNN soups prepared using ingredients having exactly the same model configuration, significantly perform better without any requirement of increasing model depth or width. For example, a 4-layer GCNII model soup prepared using 50 candidate ingredients, beats GCNII with a similar configuration on \underline{\texttt{PubMed} by $1.12 \pm 0.36$}. Our experiments across multiple GNN architectures and datasets ranging from small graphs to large graphs \texttt{\{Cora, Citeseer, PubMed, OGBN-ArXiv, OGBN-products\}} strongly validate the universal effectiveness of model soups for GNNs.      

\begin{algorithm}[tbh]
   \caption{Model soup with Graph Sampling}
   \label{alg:graphSample}
\begin{algorithmic}
   \STATE {\bfseries Input:} ingredient\_count: $I_c$; Graph $\Set{G}$, sampling\_ratio: $s$; gpu\_count: $G_c$; Hyperparameter Settings: $\Set{H}$, Sampling Criterion: $S$(node, edge, layer)
   
   \STATE candidate\_queue $\leftarrow$ $\{M_{1}, M_{2}, ..., M_{I_c}\}$ 
   \STATE soup $\leftarrow$ None
   \FOR{$i=1$ {\bfseries to} $\texttt{range}((\frac{I_c}{G_c})+1)$}
        \STATE $\{M_{1}, M_{2}, ..., M_{G_c}\} \leftarrow$ Dequeue $G_c$ ingredients  and distribute across available GPUs 
        \STATE Train all distributed $M_{i}$ in parallel with mini-batch sampling  $\texttt{sample}(\Set{G}, s, S)$ and setting $h_i \in \Set{H}$
        \STATE soup $\leftarrow$ \texttt{greedy\_weight\_interpolation} ($\{M_{1}, M_{2}, ..., M_{G_c}\} \cup$ soup)
   \ENDFOR
   \STATE Return soup
\end{algorithmic}
\label{alg:soup_graphsample}
\end{algorithm}

\subsection{Data-Centric Model soups and Large Graph Training Paradigms}
In this section, we discuss our approach for preparing data-centric model soups in scenarios when we have resource constraints to perform message passing with the entire graph data in memory, by leveraging the SOTA graph sampling and partitioning mechanisms. As MP requires nodes aggregating information from their neighbors, the integral graph structures are inevitably preserved during forward and backward propagation, thus occupying considerable running memory and time. Following Equation \ref{eq:messagepassing}, the key bottleneck relies in $\Mat{A}^{(l)}\Mat{X}^{(l)}$ for vanilla MP, requiring entire sparse adjacency matrix to be present in one GPU during training, which becomes challenging with growing topology of real-world graphs. Recently, numerous efforts with regards to graph sampling \cite{hamilton2017inductive,Zou2019LayerDependentIS,chen2018fastgcn} and partitioning \cite{chiang2019cluster,zeng2019graphsaint} have been proposed for approximating  the iterative full-batch MP to reduce the memory consumption for training within one single GPU to mitigate the aforementioned issue and scale up GNNs.

Provided the formulation of Equation \ref{eq:messagepassing}; graph sampling and partitioning paradigms pursue an optimal way to perform batch training, such that each batch will meet the memory constraint of a single GPU for message passing. We restate a unified illustration to elucidate the formulation of graph sampling and partitioning used for training our model soup ingredient $M_i$ as follows:

\vspace{-0.8cm}
\begin{multline}
    \Mat{X}^{\Mat{(K)}}_{\Set{B}_0}[M_i] = \Mat{\tilde{A}}^{\Mat{(K-1)}}_{\Set{B}_1}\sigma(\Mat{\tilde{A}}^{\Mat{(K-2)}}_{\Set{B}_2}\sigma(\cdot \cdot \cdot\sigma(\Mat{\tilde{A}}^{(0)}_{\Set{B}_K}\Mat{X}^{(0)}_{\Set{B}_K}[M_i] \\ \Mat{W}^{(0)}[M_i]  ))\cdot \cdot \cdot)\Mat{W}^{\Mat{(K-2)}}[M_i])\Mat{W}^{\Mat{(K-1)}}[M_i]
    \label{eq:modified_mp}
\end{multline}
\vspace{-0.6cm}

where $\Mat{\tilde{A}}^{\Mat{(l)}}$ indicate the adjacency matrix for the $l$-th layer sampled from the full graph, $\Set{B}_l$ is the set of nodes sampled at $l$-th layer, $M_i$ and $\Mat{W}^{(l)}[M_i]$ indicate the $i$-th candidate ingredient and weight associated with its $l$-th layer. This mini-batch training approach combined with a sampling and partitioning strategy significantly helps in alleviating  time consumption and memory usage which grow exponentially with the GNN depth. 
It is worth noting that we have explored extension to scalable infrastructure topics like distributed training with multiple GPUs to alleviate the expenses of training ingredients for model soup.

\subsubsection{Model soup with Graph Sampling} Given a large graph $\Set{G} = \{\Set{V}, \Set{E}\}$, we explore three categories of widely-used sampling strategies: \textit{node-wise sampling, edge-wise sampling, and layer-wise sampling}; to facilitate the training of candidates in our data-centric model soup. 

\vspace{-0.3cm}
\paragraph{Node-wise Sampling}: Node-wise sampling approaches propose to sample nodes from the sampling space $\Set{N}(v)$, which is 1-hop neighborhood of $v$ as $\Set{B}_{l+1}=\bigcup_{v \in \Set{B}_{l}}\{x|x \sim Q\cdot\Mat{P}_{\Set{N}(v)}\}$, where $Q$ denotes the number of samples, and $\Mat{P}$ is the sampling distribution. In our work, we have used GraphSAGE node sampling where $\Mat{P}$ is the uniform distribution. GraphSAGE sampling strategy fixes the conventional issue of ``node explosion" by clipping the neighborhood sample count to $Q$ for each node.

\vspace{-0.3cm}
\paragraph{Edge-wise Sampling}: Edge-wise sampling proposes to sample edges from the sampling space $\Set{E}(v_K)$ which is $K$-hop neighborhood of $v$ as $\Set{B}_{l+1}=\bigcup_{v \in \Set{B}_{l}}\{x|x \sim Q\cdot\Mat{P}_{\Set{E}(v_k)}\}$, where $\Set{B}_{l+1}$ denotes the set of nodes induced due to edge-sampling and $\Mat{P}$ is the uniform distribution. In our work, we have used uniform distribution for sampling a fixed amount of edges for each mini-batch training.

\vspace{-0.3cm}
\paragraph{Layer-wise Sampling}: $\Set{B}_{l+1}=\{x|x \sim Q\cdot\Mat{P}_{\Set{N}(\Set{B}_l)}\}$, where $\Set{N}(\Set{B}_l) = \bigcup_{v \in \Set{B}_l}$ denotes the 1-hop neighbourhood of all the nodes in $\Set{B}_l$ In our work, following FastGCN, the sampling distribution $\Mat{P}$ is designed using importance sampling where the probability for node $u$ of being sampled is $p(u) \propto ||\Mat{A}(u,:)||^2$. Experimentally, we found that layer-wise induced adjacency matrix is usually sparser than the others, which accounts sub-optimal performance in practice.

As shown in Algorithm \ref{alg:soup_graphsample}, for our data-centric model soup, we first initialize our soup ingredients using a common random initialization. Depending upon the resource availability, we dequeue our ingredients across different GPUs, and train them in parallel with mini-batch sampling defined by a sampling strategy (node, edge, layer) and sampling ratio. Our model soup is prepared using greedy interpolation procedure (Algorithm \ref{alg:souping}) in an ``available and added" fashion once the candidate ingredients are ready to be added to the soup.

\begin{algorithm}[tbh]
   \caption{Model soup with Graph Partitioning}
   \label{alg:graphSample}
\begin{algorithmic}
   \STATE {\bfseries Input:} ingredient\_count: $I_c$; Graph $\Set{G}$, sampling\_ratio: $s$; gpu\_count: $G_c$; Hyperparameter Settings: $\Set{H}$, Epoch Count: $E$
   \STATE Partition the graph $\Set{G}$ into fixed $K$ clusters $\Set{V} = \{V_1, V_2, ..., V_K\}$ using METIS and save.
   \STATE candidate\_queue $\leftarrow$ $\{M_{1}, M_{2}, ..., M_{I_c}\}$
   \STATE soup $\leftarrow$ None
   \FOR{$i=1$ {\bfseries to} $(\texttt{range}((\frac{I_c}{G_c})+1)$}
        \STATE $\{M_{1}, M_{2}, ..., M_{G_c}\} \leftarrow$ Dequeue $G_c$ ingredients and distribute across available GPUs 
        \FOR{For All distributed $M_{i}$ in parallel}
            \FOR{iter in \texttt{range}($E$)}
            \STATE Randomly choose $q$ clusters $\{t_1, t_2, ...,t_q\} \in \Set{V}$
            \STATE Form a subgraph $\Set{\tilde{G}}$ with nodes $\Set{\tilde{V}} = \{V_{t_1}, V_{t_2}, ..., V_{t_k}\}$ and edges $\Mat{A}_{\Set{\tilde{V}}, \Set{\tilde{V}}}$
            \STATE Train $M_{i}$ with subgraph $\Set{\tilde{G}}$ setting $h_i \in \Set{H}$
            \ENDFOR
        \ENDFOR
        \STATE soup $\leftarrow$ \texttt{greedy\_weight\_interpolation} ($\{M_{1}, M_{2}, ..., M_{G_c}\} \cup$ soup)
   \ENDFOR
   \STATE Return soup
\end{algorithmic}
\label{alg:soup_graphpartition}
\end{algorithm}

\vspace{-0.2cm}
\subsubsection{Model soup with Graph Partitioning}
Despite the success of sampling-based approaches, they still suffer from the issue of neighborhood explosion even with restricted node sampling when GNNs go deep leading to memory bottleneck. It has been observed that the efficiency of a mini-batch algorithm can be characterized by the notion of “embedding utilization”, which is proportional to the number of links between nodes in one batch or within-batch links \cite{chiang2019cluster}. Such findings motivated the design of batches using graph clustering algorithms that aim to construct partitions of nodes so that there are more graph links between nodes in the same partition than nodes in different partitions. More specifically, in graph partitioning, all the GNN layers share the same subgraph induced from the entire graph based on a partitioning strategy $\Mat{P}_{\Set{G}}$ as $\Set{B}_{K} = \Set{B}_{K-1} = ... = \Set{B}_{0} = \{x|x \sim Q\cdot\Mat{P}_{\Set{G}}\}$, such that the sampled nodes are confined in the induced subgraph. 

In our work, we propose a novel and efficient strategy for graph partition-based model soup, building upon the ClusterGCN and show that it can outperform ClusterGCN by $>1\%$ on two large graphs \texttt{OGBN-arxiv} and \texttt{Flickr}. We adopted the multi-clustering framework proposed in ClusterGCN (METIS) to partition the graph $\Set{G}$ into fixed $K$ clusters $\Set{V} = \{V_1, V_2, ..., V_K\}$ which will be used to train each candidate ingredient of the model soup. We propose to save the cluster partition to ensure that there is \textit{no overhead of expensive graph partitioning} again and again for improving the efficiency while training model soup ingredients. Once the clusters are decided, each candidate ingredient chooses $q$ (hyperparameter) vertex partitions independently and forms a subgraph that can be used for training. This partitioning strategy allows us to train comparatively deeper GNNs wrt. sampling approaches. Once the candidate ingredients are trained using hyperparameter settings $\Set{H}$, we prepare our model soup in an ``\textit{available and added}" strategy using greedy interpolation procedure (Algorithm \ref{alg:souping}) as shown in Algorithm \ref{alg:soup_graphpartition}.

\begin{figure*}[ht]
\begin{center}
\centerline{\includegraphics[width=0.95\linewidth]{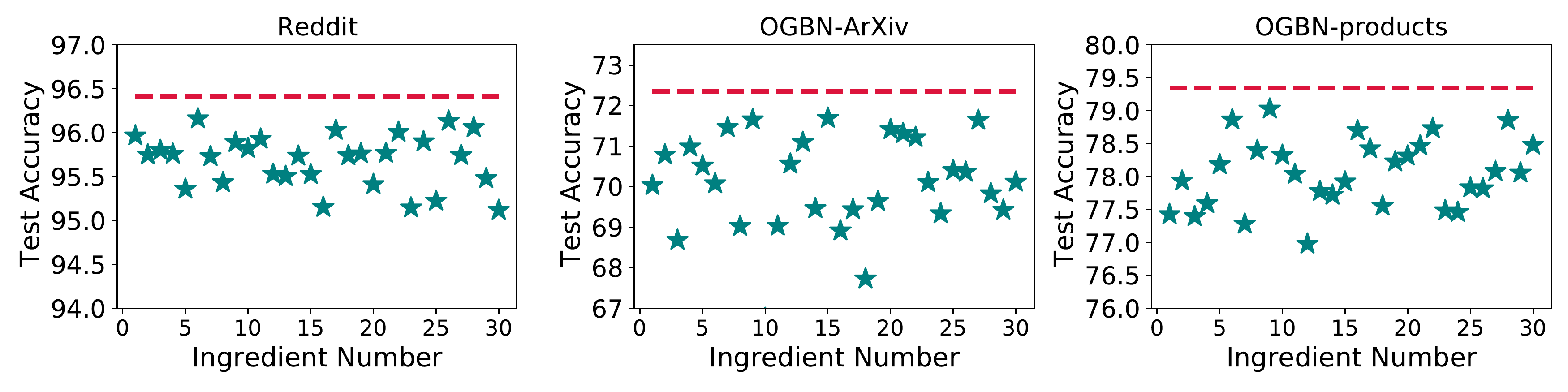}}
\vspace{-0.3cm}
\caption{Plot illustrating individual performance of each participating candidate ingredient of our data-partition soup. Red dashed line indicates the performance of the model soup generated using our greedy interpolation procedure.}
\vspace{-0.3cm}
\label{fig:orthogonal}
\end{center}
\end{figure*}

\newcolumntype{g}{>{\columncolor[gray]{.8}}c}
\begin{table*}[h]
\centering
\caption{Performance comparison of model soups generated using 50 candidate ingredients independently trained on the benchmarking datasets wrt. vanilla performance of several SOTA GNNs. Note that our ingredients have exactly the same architectural design as their vanilla counterpart. Results reported for vanilla GNNs are averaged across 30 independent runs.}
\vspace{0.1cm}
\resizebox{0.9\textwidth}{!}{\begin{tabular}{l|cg|cg|cg|cg|cg|cg}
\toprule
\multirow{2}{*}{Dataset} & \multicolumn{2}{c|}{GCN} & \multicolumn{2}{c|}{GCNII} & \multicolumn{2}{c|}{JKNet} & \multicolumn{2}{c|}{DAGNN}& \multicolumn{2}{c|}{APPNP} & \multicolumn{2}{c}{SGC}\\ 
\cmidrule{2-13}
& Vanilla & Our Soup & Vanilla & Our Soup & Vanilla & Our Soup & Vanilla & Our Soup & Vanilla & Our Soup & Vanilla & Our Soup\\
\midrule
Cora & 82.39 & 83.52 &  82.19 & 82.77 &  79.06 & 79.89 &  83.39 & 83.45 &  83.64 & 83.70 &  79.31& 80.22\\
Citeseer & 71.36  & 72.01 &  72.97 & 73.56 &  66.98  & 68.01 &  73.05 & 73.76 &  72.13 & 73.15 &  72.22& 73.53\\
PubMed & 79.56  & 80.22 &  78.06 & 79.44 &  77.24 & 77.42&  80.58 & 80.92 &  80.30 & 80.62 &  78.06& 78.94\\
ArXiv & 68.53  & 69.45 &   72.56 & 72.92 &  66.41 & 67.63 &  71.22 & 72.01 &  66.95 & 67.37 &  61.98 & 63.15\\
\bottomrule
\end{tabular}}
\vspace{-0.3em}
\label{tab:graph_soup}
\end{table*}

\vspace{-0.2cm}
\section{Experiments and Analysis}
\subsection{Dataset and Experimental Setup}
Our experiments are conducted on two GPU servers equipped with RTX A6000 and RTX 3090 GPUs. The hyper-parameters for soup ingredients corresponding to different datasets training are selected via our built-in efficient parameter sweeping functionalities from pre-defined ranges nearby our optimal setting in Appendix \ref{hyperparameter}. For our small-scale experiments, we use three standard citation network datasets: \texttt{Cora}, \texttt{Citeseer}, and \texttt{Pubmed}, while our large-scale experiments are conducted with four popular large-scale open benchmarks: \texttt{Flickr}, \texttt{Reddit}, \texttt{OGBN-ArXiv}, and \texttt{OGBN-products} (Appendix \ref{data}). For our evaluation on our chosen datasets, we have closely followed the data split settings and metrics reported by the recent benchmark \cite{duan2022a, chen2021bag}. We consider variations in the key hyperparameters  \{random seed, batch size, learning rate, weight decay, and dropout rate\} for generating the candidate ingredient models in our model soup. Unless explicitly specified, we have used 50 candidate ingredients for small-scale graphs and 30 candidate ingredients for large-scale graphs experiments and our model soup is prepared using interpolation hyperparameter $\alpha \in [0-1]$ with a step size of 0.01 in Algorithm \ref{alg:souping}. For comparison with SOTA models, we have adopted the official implementation (Appendix \ref{url_link}) of JKNet, DAGNN, APPNP, GCNII, SGC, ClusterGCN, and GraphSAINT. Note that while generating our model soup, we make sure to use exactly the same architectural design for our candidate ingredients to ensure a fair comparison. 

\begin{table*}[h]
\centering
\vspace{-0.2cm}
\caption{Illustration of the current state of various fancy architectural and regularization modifications recently proposed to facilitate deepening of GCNs and help in improving their performance. Note that  our model soup
prepared by combining the strength of 50 candidate ingredients of 2-layer GCNs can significantly outperform all these fancy methods.}
\vspace{0.1cm}
\resizebox{0.65\textwidth}{!}{\begin{tabular}{lllccccccccc} 
 \toprule
 \multirow{2}{*}{\textbf{Category}} &  \multirow{2}{*}{\textbf{Settings}} & \multicolumn{3}{c}{\textbf{Cora}} & \multicolumn{3}{c}{\textbf{Citeseer}}  & \multicolumn{3}{c}{\textbf{PubMed}} \\ 
 \cmidrule(rr){3-5}\cmidrule(rr){6-8}\cmidrule(rr){9-11}
 
 & & 2& 16 & 32 & 2& 16 & 32 & 2& 16 & 32\\
 \midrule
 Vanilla-GCN & - & 81.10 & 21.49 &21.22 &71.46 &19.59 &20.29 &79.76 &39.14 &38.77  \\
 \midrule
 Skip       & Residual& 74.73 & 20.05 & 19.57& 66.83& 20.77& 20.90& 75.27& 38.84& 38.74\\
 Connection & Initial & 79.00 & 78.61 & 78.74 &70.15 &68.41 &68.36 &77.92 &77.52 &78.18\\
            & Jumping & 80.98 & 76.04 & 75.57 & 69.33 & 58.38 & 55.03 & 77.83 & 75.62 & 75.36 \\
            & Dense   & 77.86 & 69.61 & 67.26 & 66.18 & 49.33 & 41.48 & 72.53 & 69.91 & 62.99 \\
            & Identity & 82.98 & 67.23 & 40.57 & 68.25 & 56.39 & 35.28 & 79.09 & 79.55 & 73.74 \\
 \midrule
 Normalization & BatchNorm & 69.91 & 61.20 & 29.05 & 46.27 & 26.25 & 21.82 & 67.15 & 58.00 & 55.98 \\
               & PairNorm  & 74.43 & 55.75 & 17.67 & 63.26 & 27.45 & 20.67 & 75.67 & 71.30 & 61.54 \\
               & NodeNorm  & 79.87 & 21.46 & 21.48 & 68.96 & 18.81 & 19.03 & 78.14 & 40.92 & 40.93 \\
               & CombNorm  & 80.00 & 55.64 & 21.44 & 68.59 & 18.90 & 18.53 & 78.11 & 40.93 & 40.90 \\
 \midrule
 Random Dropping & DropNode & 77.10 & 27.61 & 27.65 & 69.38 & 21.83 & 22.18 & 77.39 & 40.31 & 40.38  \\
                 & DropEdge & 79.16 & 28.00 & 27.87 & 70.26 & 22.92 & 22.92 & 78.58 & 40.61 & 40.50 \\
 
 \midrule
 \rowcolor[gray]{0.9}
 \multicolumn{2}{c}{\textbf{Our Model Soup (2-layer GCN)}} &  \multicolumn{3}{c}{\textbf{83.47 $\pm$ 0.32}} &  \multicolumn{3}{c}{\textbf{72.11 $\pm$ 0.14}}&  \multicolumn{3}{c}{\textbf{80.30 $\pm$ 0.25}}\\
 \bottomrule
\end{tabular}}
\vspace{-0.3cm}
\label{table:de-skips-comparison}
\end{table*}

\begin{table}[h]
\vspace{-0.2cm}
\caption{Performance Comaprison of GradientGCN \cite{jaiswalold} soup with 50 candidate ingredients on three WebKB datasets (Texas, Wisconsin, Cornell), and Actor dataset.}
\vspace{0.1cm}
\centering
\resizebox{0.47\textwidth}{!}{\begin{tabular}{c|cccc}
\toprule
& \textbf{Texas} & \textbf{Wisconsin}	& \textbf{Cornell} &	\textbf{Actors}\\
\midrule
GCN & 60.12$\pm$4.22 & 52.94$\pm$3.99 & 54.05$\pm$7.11 & 25.46$\pm$1.43\\
SGC & 56.41$\pm$4.25 & 51.29$\pm$6.44 & 58.57$\pm$3.44 &  26.17$\pm$1.15\\
GCNII& 64.28$\pm$2.93 &  59.19$\pm$9.07 & 58.51$\pm$1.66 &  30.95$\pm$1.04\\
JKNet & 61.08$\pm$6.23 &  52.76$\pm$5.69&  57.30$\pm$4.95   &  28.80$\pm$0.97\\
APPNP & 60.68$\pm$4.50  &  54.24$\pm$5.94  &  58.43$\pm$3.74  &  28.65$\pm$1.28\\
GradientGCN & 69.19$\pm$6.56  &  70.31$\pm$4.75 & 74.16$\pm$6.48 &  34.28$\pm$1.12\\
\midrule
\rowcolor[gray]{0.9} 
Ours & 70.45 $\pm$2.77&  72.01 $\pm$3.89& 75.90 $\pm$4.76&  34.69 $\pm$0.51\\
\bottomrule
\end{tabular}}
 \vspace{-1.4em}
\label{tab:memory_analysis}
\end{table}

\begin{table*}[h]
\caption{Performance comparison of state-of-the-art graph sampling approaches GraphSAGE, FastGCN, and LADIES wrt. our node, edge, and layer-wise sampled
soup prepared using Algorithm \ref{alg:graphSample}. Results of GraphSAGE,
FastGCN, and LADIES are reported as mean across 30 independent runs while our soup results are reported as mean of 5 independent runs, where each run uses 50 candidate ingredients.}
\label{sample-table}
\vspace{0.1cm}
\begin{center}
\resizebox{0.9\textwidth}{!}{\begin{tabular}{lcccc}
\toprule
Method & Flickr & Reddit & OGBN-ArXiv & OGBN-products \\
& N=89,250 E=899,756 & N=232,965 E=11,606,919 & N=169,343 E=1,166,243 & N= 2,449,029 E=61,859,140\\
\midrule
GraphSAGE &  53.63 $\pm$ 0.13\% & 96.50 $\pm$ 0.03\% & 71.55 $\pm$ 0.41\% & 80.61 $\pm$ 0.16\%\\
FastGCN &  49.89 $\pm$ 0.62\% & 79.50 $\pm$ 1.22\% & 66.10 $\pm$ 1.06\% & 73.46 $\pm$ 0.20\%\\
LADIES & 50.04 $\pm$ 0.39\% & 86.96 $\pm$ 0.37\% & 62.78 $\pm$ 0.89\% & 75.31 $\pm$ 0.56\%\\
\midrule
GraphSAGE Ensemble & 53.71 & 96.61 & 71.58 & 80.85\\
\midrule
\rowcolor[gray]{0.9}
Node Sampled Soup & 54.47 $\pm$ 0.13 \% & 97.28 $\pm$ 0.08 \%& 72.83 $\pm$ 0.21 \%& 81.34$\pm$ 0.28 \%\\
\rowcolor[gray]{0.9}
Edge Sampled Soup & 52.91 $\pm$ 0.56 \%& 92.66$\pm$ 0.34 \% & 70.45$\pm$ 0.29 \% & 75.58 $\pm$ 0.45 \%\\
\rowcolor[gray]{0.9}
Layer Sampled Soup & 51.08 $\pm$ 0.22 \%& 86.01$\pm$ 0.17 \% & 68.30 $\pm$ 0.54 \%& 74.95$\pm$ 0.38 \%\\
\bottomrule
\end{tabular}}
\end{center}
\label{table:data-sample}
\vspace{-0.3cm}
\end{table*}

\subsection{Model Soups and SOTA GNNs}
\label{sec:soup_gnn}
In this section, we conduct a systematic and extensive study to understand the harmonious adaptation of model soups for state-of-the-art GNNs on popular graph benchmarks \texttt{Cora}, \texttt{Citeseer}, \texttt{Pubmed}, and \texttt{OGBN-ArXiv}. 
Note that although model soups have recently attracted significant attention for large pre-trained language models, it is still unclear and unexplored if they can work for comparatively much smaller graph neural networks trained from scratch which learn from graph-structured data with relational properties unlike NLP and vision datasets having independent training samples. 
Model soups \textit{provide an orthogonal way of increasing model capacity without deepening or widening GNNs} which brings many unwanted trainability issues in GNNs.
Table \ref{tab:graph_soup} illustrates the performance summarization of our model soup generated using 50 candidate ingredients independently trained on the benchmarking datasets with respect to the vanilla performance of several SOTA GNNs following the exact same architectural design (4 layers with 256 hidden dimension) to ensure a fair comparison. The results reported for vanilla GNNs within Table \ref{tab:graph_soup} are averaged across 30 independent runs using different seed values selected at random from $[1-100]$ without replacement.

\vspace{-0.3em}
We first observe that across all datasets and GNN architecture, model soups outputs significantly outperform their vanilla counterpart. More noticeably, it improves GCN performance on \underline{Cora by $1.13\%$}, GCNII performance on \underline{PubMed by $1.38\%$}, JKNet, DAGNN, and SGC performance on \underline{OGBN-ArXiv by $1.22\%$,$0.79\%$,$1.17\%$} respectively, and APPNP performance on \underline{Citeseer by $1.02\%$}. Moreover, Table \ref{table:de-skips-comparison} illustrates the current state of various fancy architectural and regularization modifications recently proposed to facilitate deepening of GCNs and help in improving their performance. It can be clearly observed that our model soup prepared by combining the strength of 50 candidate ingredients of 2-layer GCNs can significantly outperform all these fancy methods, bolstering our claim  that model explosion by deepening and widening is necessarily not the only and right direction for building high-quality generalizable GNNs. 

\begin{table*}[h]
\vspace{-0.2cm}
\caption{Performance comparison of our Graph Partition Soup with respect to two well-known graph partitioning GNNs: ClusterGCN and GraphSAINT. Results of ClusterGCN and GraphSAINT are reported as mean across 30 independent runs while
our soup results are reported as mean of 5 independent runs, where each run uses 30 candidate ingredients.}
\label{tab:data-partition}
\begin{center}
\resizebox{0.8\textwidth}{!}{\begin{tabular}{lcccc}
\toprule
Method & Flickr & Reddit & OGBN-ArXiv & OGBN-products \\
& N=89,250 E=899,756 & N=232,965 E=11,606,919 & N=169,343 E=1,166,243 & N= 2,449,029 E=61,859,140\\
\midrule
ClusterGCN &  51.20 $\pm$ 0.13\% & 95.68 $\pm$ 0.03\% & 71.29 $\pm$ 0.44\% & 78.62 $\pm$ 0.61\%\\
GraphSAINT &  51.81 $\pm$ 0.17\% & 95.62 $\pm$ 0.05\% & 70.55 $\pm$ 0.26\% & 75.36 $\pm$ 0.34\%\\
\midrule
ClusterGCN Ensemble & 51.49 & 95.70 & 71.43 & 78.74\\
\midrule
\rowcolor[gray]{0.9}
Our Data Partition Soup & 52.23 $\pm$ 0.12 \%& 96.41$\pm$ 0.08 \%& 72.35 $\pm$ 0.19 \%& 79.34 $\pm$ 0.28 \%\\

\bottomrule
\end{tabular}}
\end{center}
\vspace{-0.3cm}
\end{table*}

\begin{table}[h]
\vspace{-0.2cm}
\caption{The memory usage of activations and the hardware throughput (higher is better) of GraphSAGE with respect to our data-centric model soup using node sampling equivalent to one-fouth of GraphSAGE on OGBN-products.}
\vspace{0.1cm}
\centering
\resizebox{0.47\textwidth}{!}{\begin{tabular}{c|ccc}
\toprule
& Act. Memory(MB) & Throughput(iter/sec) & Accuracy\\
\midrule
GraphSAGE & 415.94 & 37.69 & 79.5 $\pm$ 0.36 \\
\midrule
\rowcolor[gray]{0.9} 
Ours & 369.51 & 44.30 & 80.42 $\pm$ 0.41\\
\bottomrule
\end{tabular}}
 \vspace{-0.3cm}
\label{tab:memory_analysis}
\end{table}

\begin{table}[h]
\centering
\caption{Performance comparison of our data partition soup on OGBN-ArXiv dataset, with varying candidate ingredient counts.} 
\vspace{0.1cm}
\resizebox{0.44\textwidth}{!}{\begin{tabular}{c|ccccc}
\toprule
Ingredient Count & 10 & 20 & 30 & 50  & 100\\
\midrule
\rowcolor[gray]{0.9}
Performance & 71.426 & 71.691 & 72.353 & 72.388 & 72.814\\
\bottomrule
\end{tabular}}
\vspace{-0.3cm}
\label{tab:ingredient_count}
\end{table}

\subsection{Data-Centric Model Soup with Graph Sampling and Graph Partitioning}
\label{sec:soup_sample}
In this section, we provide experimental results for preparing data-centric model soup in scenarios when we do not have the luxury of resources to perform message passing on the entire graph, by leveraging the SOTA graph sampling (node, edge, layer) and partitioning mechanisms to prepare the candidate ingredients of the soup. Table \ref{table:data-sample} illustrates the performance comparison of state-of-the-art graph sampling approaches GraphSAGE, FastGCN, and LADIES with respect to our node, edge, and layer-wise sampled soup prepared using Algorithm \ref{alg:soup_graphsample}. Results of GraphSAGE, FastGCN, and LADIES are reported as mean across 30 independent runs while our soup results are reported as the mean of 5 independent runs. We additionally report the performance of graph ensemble prepared using the 30 independent runs of best-performing baseline (GraphSAGE). It can be clearly observed that our Node sampled soup performs best and comfortably outperforms all the baselines along with the graph ensemble which has a hefty inference cost, by significant margins. Similar to the sub-standard performance of layer-wise sampling approaches FastGCN and LADIES, our Layer sampled soup has comparatively low (although better than FastGCN and LADIES) performance, possibly because layer-wise induced adjacency matrix is usually sparser than the others, which accounts for its sub-optimal performance. 

In Table \ref{tab:memory_analysis}, we attempted to analyze the activation memory usage of activations and the hardware throughput (higher is better) of GraphSAGE (neighborhood sample size of 40) with respect to our data-centric model soup using node sampling equivalent to one-fouth of GraphSAGE (i.e., neighborhood sample size of 10) on OGBN-products. We found that with \textit{comparatively less memory requirement}, our approach can $~\sim 1\%$ better performance, eliciting the high potential in improving GNNs performance by combining the strength of multiple weak models. Next, Table \ref{tab:data-partition} illustrates the performance comparison of our Graph Partition Soup with respect to two well-known graph partitioning GNNs: ClusterGCN and GraphSAINT. Results of ClusterGCN and GraphSAINT are reported as mean across 30 independent runs while our soup results are reported as mean of 5 independent runs each generated using 30 candidate ingredients using Algorithm \ref{alg:soup_graphpartition}. We also present the performance of graph ensemble prepared using the 30 independent runs of the best-performing baseline (ClusterGCN). Across all benchmarking datasets (Flicket, Reddit, OGBN-ArXiv, and OGBN-products), our data partition soup outperforms both GraphSAINT and ClusterGCN. More noticeably, our method beats ClusterGCN with a similar architecture design by \underline{$>1\%$ margin on Flickr and OGBN-ArXiv} dataset. In addition, Figure \ref{fig:orthogonal} illustrates the individual performance of each participating candidate ingredient of our data-partition soup, and it can be clearly observed that there is \textit{orthogonal knowledge} stored in the learned weights of these networks. This gets elucidated by merging candidates and thereby improving overall performance which demonstrates the strength of our data-centric model soups.  

\subsection{Effect of Ingredient Count on Data-Centric Soup } In this section, we try to understand the strength of increasing ingredient count to our final data-centric model soup performance. Table \ref{tab:ingredient_count} illustrates the performance comparison of our data partition soup on OGBN-ArXiv dataset, with varying candidate ingredient counts. It can be clearly observed that \textit{increasing candidates generally lead to better performance}, thanks to the greedy interpolation procedure. However, due to the increase in computational and time cost, we restrict the number of ingredients to 30 for all experiments related to large graphs, and 50 for small graphs.

\subsection{Does intermediate communication benefit soup?} In this section, we attempt to answer another abaltion question: \textit{How does intermediate communication across candidate ingredients (i.e. souping at intervals during training)  benefit the performance of the final model soup?} To this end, we prepared a data partition model soup of GCNs using OGBN-ArXiv dataset, where we executed souping across candidate ingredients at regular intervals of 100 epochs during training. To our surprise, we found that the final soup performance is $-0.745\%$ less compared to our communication-free approach presumably due to a lack of diversity and orthogonal knowledge across the candidate ingredients. Moreover, intermediate communication incurs additional soup preparation overhead along with a new communication interval hyperparameter for optimization.

\section{Background Work}
Model soups \cite{wortsman2022model} proposed  a greedy mechanism for averaging weights of multiple fine-tuned large language models with varying hyperparameters uniformly. They found that averaging many
fine-tuned vision models improve out-of-domain generalization. Recently, \cite{Li2022BranchTrainMergeEP} introduced branch-train-merge which is at the intersection of model combination and distributed training. They consider the case where the training data is partitioned into different textual domains, then train an individual expert model for each domain. Merging all of these experts via weight averaging or ensembling to outperform the dense baseline of training one large model on all of the data. Lo-fi \cite{wortsman2022fi}  proposes splitting up a large language model fine-tuning job into multiple smaller jobs, and dedicating its fine-tuning across multiple nodes in isolation. Unlike standard data-parallel multi-node finetuning where gradients between nodes are communicated at each step, Lo-fi removes all communication between nodes during fine-tuning. \cite{jin2022dataless} propose a dataless knowledge fusion method and study the problem of merging individual models built on different training data sets to obtain a single model that performs well both across all data set domains and can generalize on out-of-domain data. 

Despite many recent advances in model parameter merging, to the surprise, it has not been explored for GNNs, where it can benefit the most, given the high demand for distributed and parallel training for graph data. Our work differs from recent works in model soups as: \textit{firstly}, unlike over-parameterized large language models, we explore comparatively under-parameterized GNNs; \textit{secondly}, we work with graph-structured relational data instead of independent training samples in NLP or vision; \textit{thirdly,} we explore model merging on GNNs trained from scratch rather in the fine-tuning settings. GNNs have their own set of unique training challenges and it was unexplored if the model soup mechanism will be beneficial or hurt the performance.

\subsection{Comparison to Related Work and Concurrent Ideas in Distributed GNN Training} 
Our algorithm is mainly inspired by a GNN ensembling perspective, and its aim is to optimize model accuracy with or without distributed data parallelism, by interpolating individually trained GNN model weights. Meanwhile, several recent or concurrent works have contributed to improving speed and accuracy for distributed data-parallel GNN training with efficient model averaging or randomized partitions.

\cite{ramezani2021learn} proposed a communication-efficient distributed GNN training technique (LLCG), which first trains a GNN on local machine data by ignoring the dependency between nodes among different machines, then sends the locally trained model to the server for periodic model averaging. However, LLCG heavily relies on the global Server Corrections module on the server to refine the locally learned models. CoFree-GNN \cite{cao2023communication} presents another communication-free GNN training framework, that utilizes a Vertex Cut partitioning, i.e., rather than partitioning the graph by cutting the edges between partitions, the Vertex Cut partitions the edges and duplicates the node information to preserve the graph structure.

One concurrent and independent work by \cite{zhu2023simplifying} proposed a distributed training framework that assembles independent trainers, each of which asynchronously learns a local model on locally-available parts of
the training graph. In this way, they only conducted periodic (time-based) model aggregation to synchronize the local models. Note that, unlike their periodic weight averaging of the participating trainers, our work performs averaging only after the complete training of candidates. Meanwhile, our candidates are trained by sampling different graph clusters from the input graph every epoch, to facilitate the diversity demand for model soups from the input graph. In comparison, each candidate model in \cite{zhu2023simplifying} only trains on localized subgraph assigned by randomized node/super-node partitions, without having access to entire input graph.

\vspace{-0.3em}
\section{Conclusion}
In this work, we explore a principled way to scale GNN capacity without deepening or widening which can improve its performance across multiple small and large graph.
We present a data-centric
perspective of model soups to build powerful GNNs by dividing giant graph data to build independently and parallelly trained multiple comparatively weaker GNNs without any intermediate communication, and combining their strength using a greedy interpolation soup procedure to achieve state-of-the-art performance. Moreover, we provide a wide variety of model soup preparation techniques by leveraging SOTA graph sampling and graph partitioning approaches.  Our future work will aim to develop a theoretical framework to explain the benefits of data-centric GNN soups.

\section*{Acknowledgement}

Z. Wang is in part supported by US Army Research Office Young Investigator Award W911NF2010240 and the NSF AI Institute for Foundations of Machine Learning (IFML). We also appreciate the helpful discussions with Jiong Zhu.

\bibliography{example_paper}
\bibliographystyle{icml2023}

\newpage
\appendix
\onecolumn

\section{Dataset Details}
\label{data}
Table \ref{tab:dataset_details} provided provides the detailed properties and download links for all adopted datasets. We adopt the following benchmark datasets since i) they are widely applied to develop and evaluate GNN models, especially for deep GNNs studied in this paper; ii) they contain diverse graphs from small-scale to large-scale or from homogeneous to heterogeneous; iii) they are collected from different applications including citation network, social network, etc.

\begin{table*}[h]
    \centering
    \caption{Graph datasets statistics and download links.}
    \tiny
    \resizebox{0.9\textwidth}{!}{\begin{tabular}{c@{\hspace{1\tabcolsep}}c@{\hspace{1\tabcolsep}}c@{\hspace{1\tabcolsep}}c@{\hspace{1\tabcolsep}}c@{\hspace{1\tabcolsep}}c}
      \toprule
      \textbf{Dataset} & \textbf{Nodes} & \textbf{Edges}  & \textbf{Classes}& \textbf{Download Links}\\
      \midrule
      Cora & 2,708 & 5,429   & 7 & \url{https://github.com/kimiyoung/planetoid/raw/master/data} \\
      \midrule
      Citeseer & 3,327 & 4,732   & 6 & \url{https://github.com/kimiyoung/planetoid/raw/master/data} \\
      \midrule
      PubMed & 19,717 & 44,338   & 3 & \url{https://github.com/kimiyoung/planetoid/raw/master/data} \\
      \midrule
      OGBN-ArXiv & 169,343 & 1,166,243   & 40 & \url{https://ogb.stanford.edu/}  \\
      \midrule
      Flickr & 89,250 & 899,756   & 7 & \url{PyTorch Geometic: https://arxiv.org/abs/1907.04931}  \\
      \midrule
      Reddit & 232,965 & 11,606,919 & 41 & \url{PyTorch Geometic: https://arxiv.org/abs/1706.02216}  \\
      \midrule
      OGBN-products &  2,449,029 & 61,859,140  & 47 & \url{https://ogb.stanford.edu/}  \\
       
      \bottomrule
    \end{tabular}}
    
    \label{tab:dataset_details}
\end{table*}

\section{Experimental setting of our large-scale datasets}
\label{hyperparameter}
\begin{table*}[h]
    \centering
    \caption{The searched optimal hyperparameters for all tested methods for data-centric soup.}
    \tiny
    \resizebox{0.9\textwidth}{!}{\begin{tabular}{c@{\hspace{1\tabcolsep}}c@{\hspace{1\tabcolsep}}c@{\hspace{1\tabcolsep}}c@{\hspace{1\tabcolsep}}c}
      \toprule
       Method & Flickr & Reddit & OGBN-products\\
       & Split: 0.50/0.25/0.25 & Split: 0.66 / 0.10 / 0.24 & Split: 0.10 / 0.02 / 0.88\\
       \midrule
       GraphSAGE\cite{hamilton2017inductive} & LR: 0.0001, WD: 0.0001, DP: 0.5, &  LR: 0.0001, WD: 0.0 DP: 0.2,  & LR: 0.001, WD: 0.0 DP: 0.5, \\
       & EP: 50, HD: 512, \#L: 4, BS: 1000 & EP: 50, HD: 512, \#L: 4, BS: 1000 & EP: 50, HD: 512, \#L: 4, BS: 1000\\
       \midrule
       FastGCN\cite{chen2018fastgcn} & LR: 0.001, WD: 0.0002, DP: 0.1 & LR: 0.01, WD: 0.0 DP: 0.5, & LR: 0.01, WD: 0.0 DP: 0.2\\
       & EP: 50, HD: 512, \#L: 2, BS: 5000 & EP: 50, HD: 256, \#L: 2, BS: 5000 & EP: 50, HD: 256, \#L: 2, BS: 5000\\
       \midrule
       LADIES\cite{Zou2019LayerDependentIS}& LR: 0.001, WD: 0.0002, DP: 0.1, & LR: 0.01, WD: 0.0001 DP: 0.2, & LR: 0.01, WD: 0.0 DP: 0.2\\
       & EP: 50, HD: 512, \#L: 2, BS: 5000 & EP: 50, HD: 512, \#L: 2, BS: 5000 & EP: 30, HD: 256, \#L: 2, BS: 5000\\
       \midrule
       ClusterGCN\cite{chiang2019cluster} & LR: 0.001, WD: 0.0002, DP: 0.2, & LR: 0.0001, WD: 0.0 DP: 0.5 & LR: 0.001, WD: 0.0001 DP: 0.2,\\
       & EP: 30, HD: 256, \#L: 2, BS: 5000& EP: 50, HD: 256, \#L: 4, BS: 2000 & EP: 40, HD: 128, \#L: 4, BS: 2000 \\
       \midrule
       GraphSAINT\cite{zeng2019graphsaint} & LR: 0.001, WD: 0.0004, DP: 0.2 & LR: 0.01, WD: 0.0002 DP: 0.7 & LR: 0.01, WD: 0.0 DP: 0.2,\\
       & EP: 50, HD: 512, \#L: 4, BS: 5000 & EP: 30, HD: 128, \#L: 2, BS: 5000 & EP: 40, HD: 128, \#L: 2, BS: 5000 \\
      \bottomrule
    \end{tabular}}
    \label{tab:dataset_details}
\end{table*}

\section{Code adaptation URL for our baselines}
\label{url_link}
\begin{table*}[h]
    \centering
    \caption{Method and their official implementation used in our work.}
    \tiny
    \resizebox{0.6\textwidth}{!}{\begin{tabular}{l@{\hspace{1\tabcolsep}}l@{\hspace{1\tabcolsep}}}
      \toprule
       Method & Download URL\\
       \midrule
       JKNet\cite{xu2018representation} & \url{https://github.com/mori97/JKNet-dgl}\\
       DAGNN\cite{yang2021dagnn} & \url{https://github.com/vthost/DAGNN}\\
       APPNP\cite{Klicpera2019PredictTP} & \url{https://github.com/gasteigerjo/ppnp}\\
       GCNII\cite{Chen2020SimpleAD} & \url{https://github.com/chennnM/GCNII}\\
       SGC\cite{wu2019simplifying} & \url{https://github.com/Tiiiger/SGC}\\
       ClusterGCN\cite{chiang2019cluster} & \url{https://github.com/benedekrozemberczki/ClusterGCN}\\
       GraphSAINT\cite{zeng2019graphsaint} & \url{https://github.com/GraphSAINT/GraphSAINT}\\
      \bottomrule
    \end{tabular}}
    \label{tab:dataset_details}
\end{table*}


\end{document}